\documentclass[runningheads]{llncs}
\usepackage{graphicx}

\usepackage{tikz}
\usepackage{comment}
\usepackage{amsmath,amssymb} 
\usepackage{color}
\usepackage{overpic}
\usepackage{multirow}
\usepackage{makecell}

\makeatletter
\newcommand*{\defeq}{\mathrel{\rlap{%
                     \raisebox{0.3ex}{$\m@th\cdot$}}%
                     \raisebox{-0.3ex}{$\m@th\cdot$}}%
                     =}
\makeatother

\newcommand{\fgrid}[0]{\mathbf{F}}
\newcommand{\point}[0]{\mathbf{p}}

\begin{document}
\pagestyle{headings}
\mainmatter
\def\ECCVSubNumber{7}  

\title{Implicit Feature Networks for Texture Completion from Partial 3D Data} 
\titlerunning{IF-Net for Texture Completion}

\authorrunning{Chibane et al.}
\author{Julian Chibane \and
Gerard Pons-Moll}
\institute{Max Planck Institute for Informatics \\ \email{\{jchibane,gpons\}@mpi-inf.mpg.de}} 

\maketitle

\begin{abstract}
Prior work to infer 3D texture use either texture atlases, which require uv-mappings and hence have discontinuities, or colored voxels, which are memory inefficient and limited in resolution. Recent work, predicts RGB color at every XYZ coordinate forming a texture field, but focus on completing texture given a single 2D image.
Instead, we focus on 3D \emph{texture} and geometry completion from partial and incomplete \textit{3D scans}. 
IF-Nets~\cite{chibane20ifnet} have recently achieved state-of-the-art results on 3D geometry completion using a multi-scale deep feature encoding, but the outputs lack texture. In this work, we generalize IF-Nets to texture completion from partial textured scans of humans and arbitrary objects. 
Our key insight is that 3D texture completion benefits from incorporating local and global deep features extracted from \emph{both} the 3D partial texture and completed geometry. 
Specifically, given the partial 3D texture and the 3D geometry completed with IF-Nets, our model successfully in-paints the missing texture parts in consistence with the completed geometry. Our model won the SHARP ECCV'20 challenge, achieving highest performance on all challenges.

\end{abstract}

\section{Introduction}
\label{sec:intro}
Motivated by difficulties in the 3D capturing process of complete scans the ECCV 2020 SHApe Recovery from Partial textured 3D scans (SHARP) Workshop\footnote{https://cvi2.uni.lu/sharp2020/} initialized a challenge on reconstructing full 3D textured meshes from partial 3D scan acquisitions. Complete 3D scanning, amongst other challenges, need to capture varying levels of details from the scene, need to handle occlusions, transparency, movement and require bulky studio setups. On the other hand, mobile hand-held scanners\footnote{Artec3D, https://www.artec3d.com} allow an adaptive usage. Longer acquisition times, can result in partial scans with inaccuracies which we target to complete in this work.

For the geometry reconstruction we rely on the state-of-the-art Implicit Feature Networks~\cite{chibane20ifnet} (IF-Nets) neural 3D processing architecture. We find exciting results for reconstruction of humans: even in the absence of whole body parts, as arms or feet, in the corrupted input, the fully learned IF-Nets are generating plausible human shape completions, we expected to require a human body model.

However, the reconstructions are missing texture - an integral part of a human scan, consumers are interested in. There are two common texture representation which both have limitations: 1) texture can be represented as 2D texture atlases stored as images, which need geometry templates to find uv-mappings from 3D to 2D and have hard to handle discontinuities and 2) colorized voxel grids, which are memory inefficient and restrict to low-frequency texture. A novel texture representation, learned texture fields, predict rgb color at a specific xyz coordinate. However, prior work do inference only based on 2D image evidence \cite{Niemeyer2020CVPR,Oechsle2019ICCV,pifuSHNMKL19,i_mildenhall2020nerf}. In this work we extend the IF-Net architecture for use of texture completion from colored, partial 3D data: humans and arbitrary objects.
As input to our texture field prediction we use the partial textured input surface and an IF-Net untextured geometry completion, and feed both as voxelized grids into our network. For every point on the input full geometry we predict the rgb color. We find our model successfully inpaints the missing texture parts in consistence with the completed geometry.

\section{Related Work}
\label{sec:related}
\paragraph{Implicit Function Leaning (IFL)}
IFL methods use either binary signed distance functions~\cite{i_DeepSDF,i_IMGAN19,i_DeepLevelSet,i_jiang2020local} or occupancies~\cite{i_OccNet19,i_genova2019deep,chibane20ifnet,pifuSHNMKL19} as shape representation for learning. 
These methods incorporate a simple but very powerful trick: a network predicts occupancy or the SDF valuea at continuous point locations ($x$-$y$-$z$). \\
IF-Nets\cite{chibane20ifnet} showed state-of-the-art performance on reconstruction from partial 3D. However, the reconstructions are missing texture. We therefor, in this work, extend IF-Nets to reconstruction of complete, textured reconstructions. \\
Another application of IFL methods is novel view synthesis, that is, generating novel view-points given one or multiple images~\cite{i_sitzmann2019scene,i_mildenhall2020nerf}, however, this is not the focus of our work.

\paragraph{Texture fields}
Recent work, predicts RGB color at every XYZ coordinate forming a texture field, but focus on completing texture given a single 2D image \cite{pifuSHNMKL19,Oechsle2019ICCV} or multiple 2D images \cite{i_mildenhall2020nerf,Niemeyer2020CVPR}.
Instead, we focus on 3D \emph{texture} and geometry completion from partial and incomplete \textit{3D scans}, which is required for the given challenge.

\section{Method}
\label{sec:method}
Our goal is to reconstruct a completed textured surface of humans and objects given only a textured partial observation. 
To predict the full, textured reconstruction 
we extend IF-Nets~\cite{chibane20ifnet} to inference of texture fields. (Sec. \ref{subsec:method.texture_inference})
To obtain geometry reconstructions we train a regular IF-Net (Sec. \ref{subsec:method.geometry_reconstruction}).

\begin{figure}[t]

\begin{overpic}[width=1\linewidth]{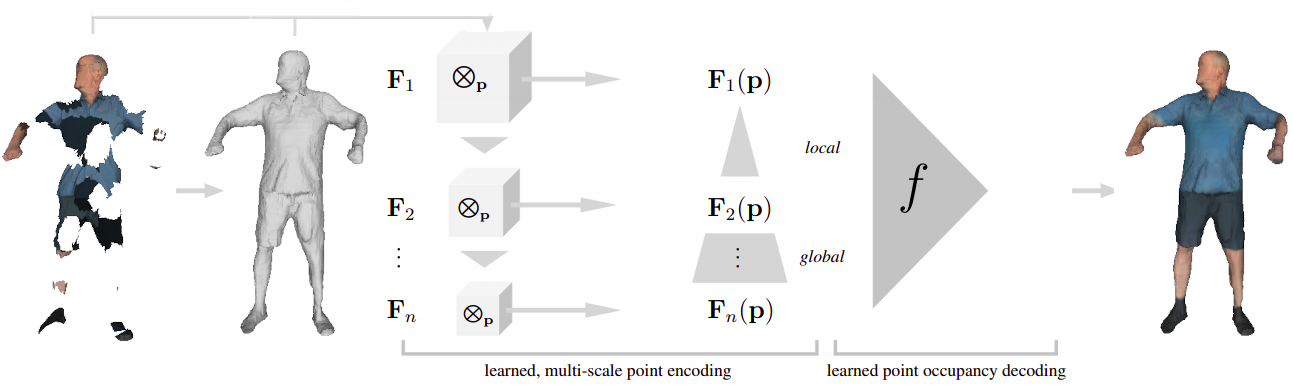} 
    \put(5,1){\tiny Input}
    \put(12,13){\tiny IF-Net}
    \put(44,1){\begin{minipage}{\textwidth}   \begin{center}\tiny Full \\ Reconstruction
        \end{center}\end{minipage} }
    \put(-27,1){\begin{minipage}{\textwidth}   \begin{center}
        \tiny Geometry \\ Reconstruction \end{center} \end{minipage} }
    \put(77,14.5){\scriptsize $\begin{bmatrix}
R\\
G\\
B
\end{bmatrix}$}

\end{overpic}

\caption{Overview of inference-time prediction of textured, complete surfaces from corrupted, partial input. We train a standard IF-Net to reconstruct untextured geometry given the corrupted, partial input. 
An rgb value at point $\point$ is predicted by explicitly extracting local and global deep features computed from \emph{both} the 3D partial texture and completed geometry. Our key insight is that this enables our model to generate texture in consistence with the completed geometry.
}
\label{fig:method_overview}
\end{figure}
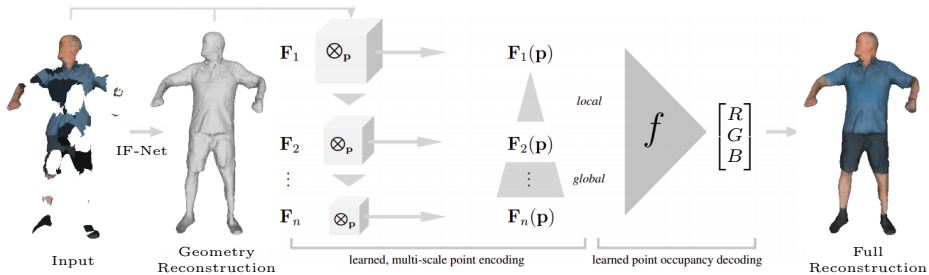

\subsection{Texture Inference}
\label{subsec:method.texture_inference}

We learn to complete the texture of a given partial 3D surface, $\mathcal{T}_i$. For this, we learn a texture field, that is, a mapping of points $\mathbf{p}$, on a complete but untextured surface, to a color at that point. As input $\mathbf{x}$ to the texture field prediction, we additionally use a complete untextured surface geometry of the object. During training the complete surface geometry is provided by the ground truth surfaces, $\mathcal{S}_i$, during inference it is predicted as described in Sec \ref{subsec:method.geometry_reconstruction}.

\paragraph{Encoding}
We construct an input encoding \textit{aligned} with the input data. That is, using a 3D CNN we neurally process the input to feature grids, that lie in the same Euclidean space as the input. 
By recursively applying $n$ times, 3D convolutions followed by down scaling through max pooling, we create \emph{multi-scale deep feature grids} $\fgrid_1,..,\fgrid_n$. The feature grids at the early stages capture texture and shape details, whereas feature grids at the late stages have large receptive fields and capture global structure, for example to segment garments for giving them a coherent texture. See \cite{chibane20ifnet} for details.\\
To adapt the procedure for texture inference, we feed 4 channels of input data into the encoder, 3 channels for the textured partial observation - one channel per rgb - and the fourth channel encodes the complete surface geometry.
For usage with a 3D CNN we need to voxelize the surfaces, we do this by densely sampling surface points and marking their nearest neighbor in an enclosing voxel grid. For the partial textured input, we mark the nearest neighbour voxels by the rgb intensities in $\{0,\dots,255\}$ and all other with $-1$. For the complete surface geometry, we mark voxels selected as nearest neighbours with 1 and all other with 0.
We denote the encoder as $g(\mathbf{x}) \defeq \fgrid_1,..,\fgrid_n$, where our input is denoted as $\mathbf{x}$.

\paragraph{Decoding}
We create a point $\point$ specific encoding $\fgrid_1(\point),..,\fgrid_n(\point)$ by extracting deep features at the location $\point$ from the global encoding $g(\mathbf{x}) = \fgrid_1,..,\fgrid_n$. This is possible because $g(\mathbf{x})$ is kept aligned with Euclidean space. Since our feature grids are discrete, we use trilinear interpolation to query continuous 3D points $\point \in \mathbb{R}^3$. See \cite{chibane20ifnet} for details.\\
The point encoding $\fgrid_1(\point),..,\fgrid_n(\point)$, is then fed into a point-wise decoder $f(\cdot)$, parameterized by a fully connected neural network with ReLU activations, to regress the rgb color at $\point$.

\paragraph{Training}
At training time we use the textured partial surfaces, $\mathcal{T}_i$. and ground truth complete surface, $\mathcal{S}_i$, without texture, as inputs for the network. The task of the network is then, to learn to regress the correct color $\mathrm{rgb(\point,\mathcal{S}_i)}$ at a surface point $\point \in \mathcal{S}_i$ of the ground truth surface. We construct training data by sampling points $\point$ uniform at random on the surface $\mathcal{S}_i$ and find their ground truth $\mathrm{rgb(\point,\mathcal{S}_i)}$ values. More specifically, we have $\mathcal{S}_i$ available as mesh .obj-files with attached texture atlases. Therefore, to find $\mathrm{rgb(\point,\mathcal{S}_i)}$, we check for the triangle $\point$ is on, find the uv-coordinates of $\point$ by Barycentric interpolation of the uv-coordinates of the triangle vertices and extract the rgb value from the texture atlas at the found uv location. We use an L1 mini-batch loss:

\begin{equation}
\small
\mathcal{L}_\mathcal{B}({\mathbf{w}}) 
\defeq \sum_{\mathbf{x} \in \mathcal{B}} \sum_{\point \in \mathcal{P}} ||f^{\mathbf{w}}(g_{\mathbf{w}}(\mathbf{x},\point)) - \mathrm{rgb}(\point,\mathcal{S}_{\mathbf{x}})||_1\\
 \label{eq:loss}
\end{equation} 
where $\mathcal{B}$ is a input mini-batch, $\mathcal{P}$ is a sub-sample of points, $\mathbf{w}$ are the neural parameters of encoder and decoder, and 
$g_{\mathbf{w}}(\mathbf{x},\point) \defeq \fgrid_1^\mathbf{w}(\point),\hdots, \fgrid_n^\mathbf{w}(\point)$. 

\paragraph{Inference} At inference time we feed the partial textured surface, $\mathcal{T}_i$, and its surface reconstruction (untextured) (Sec. \ref{subsec:method.geometry_reconstruction}) as input into the texture inference.
The surface reconstruction is a mesh and for all its vertices we regress the rgb value and attach it to the mesh. This yields our final colored surface reconstruction. See Figure \ref{fig:method_overview} for an overview of our method.

\subsection{Geometry Reconstruction}
\label{subsec:method.geometry_reconstruction}
During inference, we predict the geometry using an IF-Net and condition the texture generation (Sec. \ref{subsec:method.texture_inference}) on this geometry. 
For learning the geometry reconstruction, we use a standard IF-Net~\cite{chibane20ifnet}. Our training data consists of pairs $(\mathcal{T}_i,\mathcal{S}_i)_{i}$, of partial surfaces, $\mathcal{T}_i$, and complete ground truth surfaces, $\mathcal{S}_i$. 
The shape prediction given $\mathcal{T}_i$, of IF-Nets is done implicitly, by predicting for any point $\mathbf{p}$ in the continuous space $\mathbb{R}^3$, if $\mathbf{p}$ is inside (classification as $1$) or outside (classification as $0$) of the ground truth $\mathcal{S}_i$. For inference of surfaces, these predictions are evaluated on a dense gird in $\mathbb{R}^3$ and converted to a mesh using marching cubes~\cite{c_marching_cubes}.

\paragraph{Encoding \& decoding} is done with the same architecture described above. However, the input to the encoder consists \textit{only} of one (untextured) channel, created from the partial surface. To encode the surface we again, sample points on it and create a voxel grid. We mark a voxel with 1, if it is a nearest neighbour of a point, and 0 else. We use a grid resolution of $256^3$. Instead of 3 values for rgb we decode into one value for the above mentioned occupancy classification.

\paragraph{Training} For each ground truth surface, $S_i$, we sample 100.000 points and compute their occupancy. Samples are created in the surface vicinity to concentrate the model capacity on details. To this end, we first uniformly at random sample surface points $\point_\mathcal{S}$ and apply Gaussian distributed displacements $\mathbf{n} \sim \mathcal{N}(0, \boldsymbol{\Sigma})$, i.e. $\point \defeq \point_\mathcal{S} + \mathbf{n}$. We use a diagonal co-variance matrix $\boldsymbol{\Sigma} \in \mathbb{R}^{3\times{3}}$ with entries $\boldsymbol{\Sigma}_{i,i} = \sigma$. To capture detail we sample $50\%$ of the point samples very near the surface with a small $\sigma_1 = 0.015$, and to learn about the global, more homogeneously filled space, we sample $50\%$ with a larger $\sigma_2 = 0.2$. During training 50.000 points are sub-sampled and learning is unaltered from \cite{chibane20ifnet}.

\section{SHARP ECCV 2020 Challenge}
\label{sec:challenge}
In this section we present the challenges and the qualitative results of our method. For a quantitative result analysis, including other participants and baselines, we refer to the SHARP summary paper corresponding to this workshop.

\subsection{Challenge 1}

\subsubsection{Track 1}
\begin{figure}[t]
\begin{overpic}[width=1\linewidth]{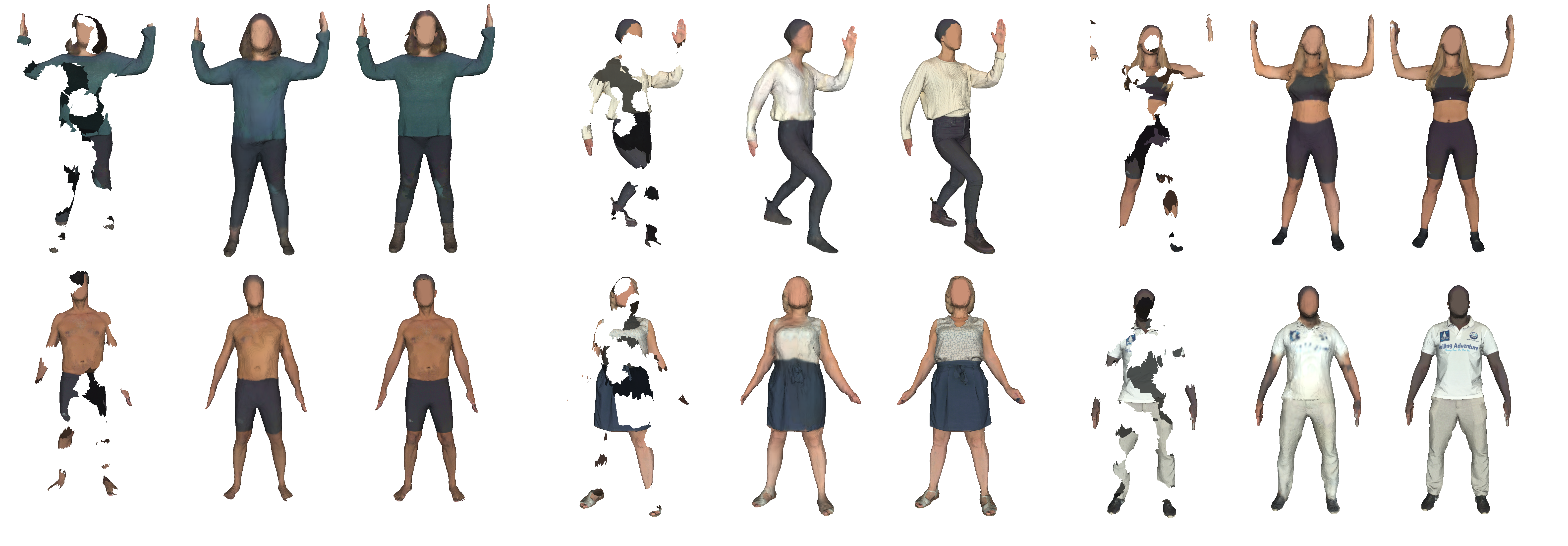} 
    \put(0,18){\tiny a)}
    \put(35,18){\tiny b)}
    \put(68,18){\tiny c)}

    \put(0,1){\tiny d)}
    \put(35,1){\tiny e)}
    \put(68,1){\tiny f)}

\end{overpic}

\caption{Six testing results a)-f) from our method (unseen during training), left to right: textured, partial input, textured reconstruction, ground truth.  Our geometry reconstructions notably show generated body parts in correct positions even for largely missing body parts, see right arm of person e). Our texture generation is visibly coherent with the reconstructed geometry, through the geometry conditioning, see transition from shirt to leggings of b). The texture generation learned can rely on symmetry for generation, see the right foot of person in c).
}
\label{fig:result_track1}
\end{figure}
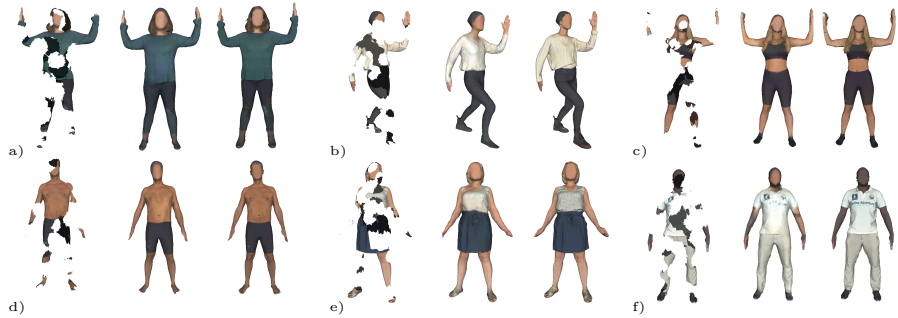

For track 1 we were kindly given access to the 3DBodyTex.v2 dataset, similar in quality to 3DBodyTex~\cite{3DBodyTex}\footnote{https://cvi2.uni.lu/datasets/}. It consists of about 2500 of human scans with high diversity in  poses. Some humans wear minimal fitness clothing others are fully clothed in varying clothing. We train our algorithm solely on the officially provided training set. 
The task is to reconstruct a completed textured 3D surface from a 3D human scan with large missing areas. The ground truth surfaces, $\mathcal{S}_i$, are the raw scans. The textured partial surfaces, $\mathcal{T}_i$, are generated synthetically from the ground truth shapes, using the provided script with default values\footnote{https://gitlab.uni.lu/cvi2/eccv2020-sharp-workshop/-/tree/master}. For each ground truth mesh we generate 4 partial scans with different incompleteness patterns. The partial data should simulate an acquisition produced by a hand-held 3D scanner. For all scans we found the bounding box with $$(min_x, max_x, min_y, max_y,, min_z, max_z) = (-0.8, 0.8, -0.15, 2.1, -0.8, 0.8)$$ to be sufficient and use it to place our voxel grid encoding (see Sec. \ref{subsec:method.texture_inference}). 
In Fig. \ref{fig:result_track1} we show the results obtained with our method from Sec. \ref{sec:method}.

\subsubsection{Track 2}
\begin{figure}[t]
\begin{overpic}[width=1\linewidth]{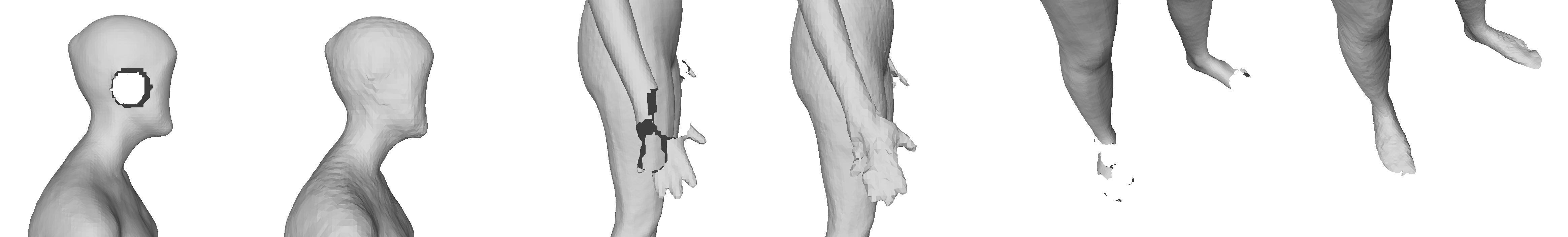} 
    \put(0,14){\tiny a)}
    \put(33,14){\tiny b)}
    \put(63,14){\tiny c)}

\end{overpic}

\caption{Results on evaluation data (GT disclosed), for reconstruction of the ear region a), hand b) and feet c) regions. Although trained for geometry reconstruction of holes at arbitrary regions, plausible shape completion for these regions is achieved.}
\label{fig:result_track2}
\end{figure}

In Track 2, the task is to reconstruct fine geometry details of the body such as ears, fingers, feet or a nose, without texture. For this we are given around 1000 detailed ground truth meshes, $\mathcal{S}_i$, obtained by fitting a human body model to human data in minimal clothing from 3DBodyTex.v2. See \cite{8803819,AutomaticFitting} for methods on automatic and robust model fitting proposed by the dataset authors. On these ground truth meshes a scanning process of the Shapify booth is simulated in software. Additionally, we again shoot holes into the mesh using the provided script in its default setting, shooting holes also at other body parts. Again, we generate 4 partial scans with different incompleteness patterns. Only when the evaluation data was released, we found, that holes only occur at ears, fingers, feet or noses. Therefore, an obvious enhancement for our training would be to concentrate model capacity to those regions by only shooting holes there. We reuse the bounding box found for Track 1.
In Fig. \ref{fig:result_track2} we show qualiatative results obtained with our method, IF-Nets.

\subsection{Challenge 2}
\begin{figure}[t]
\includegraphics[width=1\linewidth]{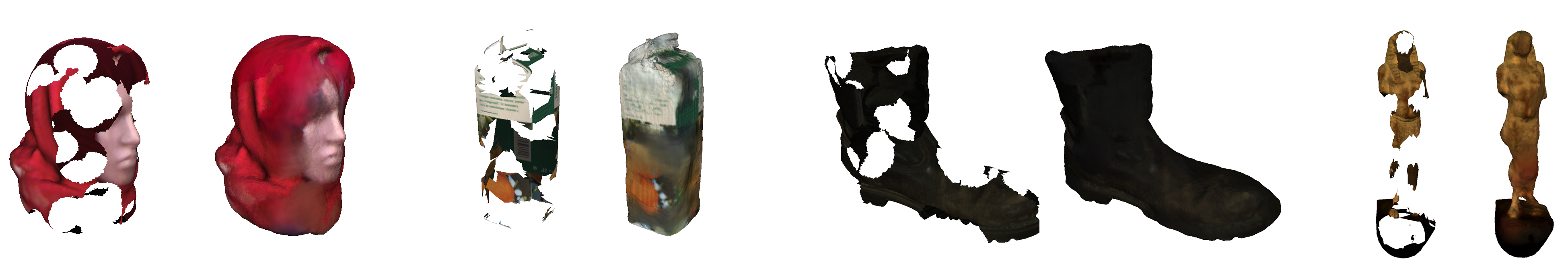}

\caption{Results on challenge evaluation data (GT disclosed), for reconstruction of textured, arbitrary objects. This is a highly challenging dataset and problem: objects come from no common class. Thus, no shape prior of objects can be learned - the textured reconstruction can solely rely on surface statistics common in any object. Our network performs reasonable textured reconstruction of what we believe to be a head, orange juice, a shoe and an Egyptian statue.}
\label{fig:result_challenge2}
\end{figure}

In this challenge we are given a dataset (3DObjectTex.v1) of 2000, high quality, textured 3D ground truth scans, $\mathcal{S}_i$, of very diverse objects (real and stuffed animals, statues, technology, furniture,...). When scanning with a hand-held 3D scanner, technical infeasibility of reaching some parts of an object can lead to partial textured scans. Therefore, the task of this challenge is to reconstruct a textured, complete mesh from such partial data. In contrast to challenge 1, the diversity of objects makes it infeasible to build object specify shape priors. As before, we create 4 partial scans per ground truth. Due to huge size variations, we rescale all objects to lie in [-0.5,0.5] in each dimension, by subtracting each vertex with the object center and multiplying by the inverse of its longest bounding box edge.
In Fig. \ref{fig:result_challenge2} we show qualitative results obtained with our IF-Nets.

\section{Discussion and Conclusions}
\label{sec:conclusions}
We proposed an extension of IF-Nets for \textit{textured}, complete 3D reconstruction from textured, highly incomplete 3D scans. 
For this we predict a continuous texture field, that is, we predict the rgb color at any given point in 3D. Our model won the SHARP ECCV'20 challenge, achieving highest performance on all challenges.

We found that 3D texture completion benefits from incorporating local and global deep features extracted from \emph{both} the 3D partial texture and completed geometry. 
Specifically, given the partial 3D texture and the 3D geometry completed with IF-Nets, our model successfully in-paints plausible texture in consistence with the completed geometry. We hypothesize, this is because areas of homogeneous texture are often strongly correlated with the geometry: a uni-colored t-shirt or dress can be well textured given a geometry segmentation.

Future work can address to generate higher frequency patterns of texture currently missing in the completions. A promising direction is to use our 3D texture completions, in a coarse to fine framework, as a conditioning for high frequency generative models in 2D.

%
%
\bibliographystyle{splncs04}
\bibliography{egbib}

\begin{thebibliography}{10}
\providecommand{\url}[1]{\texttt{#1}}
\providecommand{\urlprefix}{URL }
\providecommand{\doi}[1]{https://doi.org/#1}

\bibitem{i_IMGAN19}
Chen, Z., Zhang, H.: Learning implicit fields for generative shape modeling.
  In: {IEEE} Conference on Computer Vision and Pattern Recognition, {CVPR}
  2019, Long Beach, CA, USA, June 16-20, 2019. pp. 5939--5948 (2019)

\bibitem{chibane20ifnet}
Chibane, J., Alldieck, T., Pons-Moll, G.: Implicit functions in feature space
  for 3d shape reconstruction and completion. In: {IEEE} Conference on Computer
  Vision and Pattern Recognition (CVPR). {IEEE} (jun 2020)

\bibitem{i_genova2019deep}
Genova, K., Cole, F., Sud, A., Sarna, A., Funkhouser, T.: Deep structured
  implicit functions. In: 2020 {IEEE} Conference on Computer Vision and Pattern
  Recognition (2019)

\bibitem{i_jiang2020local}
Jiang, C.M., Sud, A., Makadia, A., Huang, J., Nie{\ss}ner, M., Funkhouser, T.:
  Local implicit grid representations for 3d scenes. In: 2020 {IEEE} Conference
  on Computer Vision and Pattern Recognition (2020)

\bibitem{c_marching_cubes}
Lorensen, W.E., Cline, H.E.: Marching cubes: {A} high resolution 3d surface
  construction algorithm. In: Computer Graphics and Interactive Techniques. pp.
  163--169 (1987). \doi{10.1145/37401.37422},
  \url{https://doi.org/10.1145/37401.37422}

\bibitem{i_OccNet19}
Mescheder, L.M., Oechsle, M., Niemeyer, M., Nowozin, S., Geiger, A.: Occupancy
  networks: Learning 3d reconstruction in function space. In: {IEEE} Conference
  on Computer Vision and Pattern Recognition, {CVPR} 2019, Long Beach, CA, USA,
  June 16-20, 2019. pp. 4460--4470 (2019)

\bibitem{i_DeepLevelSet}
Michalkiewicz, M., Pontes, J.K., Jack, D., Baktashmotlagh, M., Eriksson, A.P.:
  Deep level sets: Implicit surface representations for 3d shape inference.
  CoRR  \textbf{abs/1901.06802} (2019), \url{http://arxiv.org/abs/1901.06802}

\bibitem{i_mildenhall2020nerf}
Mildenhall, B., Srinivasan, P.P., Tancik, M., Barron, J.T., Ramamoorthi, R.,
  Ng, R.: Nerf: Representing scenes as neural radiance fields for view
  synthesis. arXiv preprint arXiv:2003.08934  (2020)

\bibitem{Niemeyer2020CVPR}
Niemeyer, M., Mescheder, L., Oechsle, M., Geiger, A.: Differentiable volumetric
  rendering: Learning implicit 3d representations without 3d supervision. In:
  Proceedings IEEE Conf. on Computer Vision and Pattern Recognition (CVPR)
  (2020)

\bibitem{Oechsle2019ICCV}
Oechsle, M., Mescheder, L., Niemeyer, M., Strauss, T., Geiger, A.: Texture
  fields: Learning texture representations in function space. In: International
  Conference on Computer Vision (Oct 2019)

\bibitem{i_DeepSDF}
Park, J.J., Florence, P., Straub, J., Newcombe, R.A., Lovegrove, S.: Deepsdf:
  Learning continuous signed distance functions for shape representation. In:
  {IEEE} Conference on Computer Vision and Pattern Recognition, {CVPR} 2019,
  Long Beach, CA, USA, June 16-20, 2019. pp. 165--174 (2019)

\bibitem{3DBodyTex}
{Saint}, A., {Ahmed}, E., {Shabayek}, A.E.R., {Cherenkova}, K., {Gusev}, G.,
  {Aouada}, D., {Ottersten}, B.: 3dbodytex: Textured 3d body dataset. In: 2018
  International Conference on 3D Vision (3DV). pp. 495--504 (2018)

\bibitem{8803819}
{Saint}, A., {Rahman Shabayek}, A.E., {Cherenkova}, K., {Gusev}, G., {Aouada},
  D., {Ottersten}, B.: Bodyfitr: Robust automatic 3d human body fitting. In:
  2019 IEEE International Conference on Image Processing (ICIP). pp. 484--488
  (2019)

\bibitem{AutomaticFitting}
Saint, A., Shabayek, A., Aouada, D., Ottersten, B., Cherenkova, K., Gusev, G.:
  Towards automatic human body model fitting to a 3d scan. In: 8th
  International Conference and Exhibition on 3D Body Scanning and Processing
  Technologies, Montreal QC, Canada. pp. 274--280 (10 2017).
  \doi{10.15221/17.274}

\bibitem{pifuSHNMKL19}
Saito, S., , Huang, Z., Natsume, R., Morishima, S., Kanazawa, A., Li, H.: Pifu:
  Pixel-aligned implicit function for high-resolution clothed human
  digitization. arXiv preprint arXiv:1905.05172  (2019)

\bibitem{i_sitzmann2019scene}
Sitzmann, V., Zollh{\"o}fer, M., Wetzstein, G.: Scene representation networks:
  Continuous 3d-structure-aware neural scene representations. In: Advances in
  Neural Information Processing Systems. pp. 1119--1130 (2019)

\end{thebibliography}
\end{document}